\documentclass[10pt,twocolumn,letterpaper]{article}

\usepackage{wacv}
\usepackage{times}
\usepackage{epsfig}
\usepackage{graphicx}
\usepackage{amsmath}
\usepackage{amssymb}
\usepackage{booktabs}
\usepackage{pdfpages}

\usepackage{xurl}
\usepackage{comment}
\usepackage{amsmath,amssymb} 
\usepackage{color}
\usepackage{subcaption}
\usepackage{mathtools}
\usepackage{caption}
\usepackage{fancyvrb} 
\VerbatimFootnotes    

\def\wacvPaperID{174} 

\wacvalgorithmstrack   

\wacvfinalcopy 

\ifwacvfinal
\usepackage[breaklinks=true,bookmarks=false]{hyperref}
\else
\usepackage[pagebackref=true,breaklinks=true,colorlinks,bookmarks=false]{hyperref}
\fi

\usepackage[capitalize]{cleveref}
\begin{document}



\newif\ifdraft
\draftfalse
\drafttrue 

\definecolor{orange}{rgb}{1,0.5,0}
\definecolor{violet}{RGB}{70,0,170}
\definecolor{magenta}{RGB}{170,0,170}
\definecolor{dgreen}{RGB}{0,150,0}

\ifdraft
 \newcommand{\PF}[1]{{\color{red}{\bf PF: #1}}}
 \newcommand{\pf}[1]{{\color{red} #1}}
 \newcommand{\WL}[1]{{\color{blue}{\bf WL: #1}}}
 \newcommand{\wl}[1]{{\color{blue} #1}}
  \newcommand{\ME}[1]{{\color{dgreen}{\bf ME: #1}}}
 \newcommand{\me}[1]{{\color{dgreen} #1}}
\else
 \newcommand{\PF}[1]{}
 \newcommand{\pf}[1]{#1}
 \newcommand{\WL}[1]{}
 \newcommand{\wl}[1]{#1}
 \newcommand{\ME}[1]{}
  \newcommand{\me}[1]{#1}
\fi

\newcommand{\parag}[1]{\vspace{-3mm}\paragraph{#1}}
\newcommand{\sparag}[1]{\subparagraph{#1}}
\renewcommand{\floatpagefraction}{.99}

\newcommand{\bA}{\mathbf{A}}
\newcommand{\bC}{\mathbf{C}}
\newcommand{\bD}{\mathbf{D}}
\newcommand{\bI}{\mathbf{I}}
\newcommand{\bH}{\mathbf{H}}
\newcommand{\bP}{\mathbf{P}}
\newcommand{\bR}{\mathbf{R}}
\newcommand{\bZ}{\mathbf{Z}}

\newcommand{\real}{\mathbb{R}}

\newcommand{\bc}{\mathbf{c}}
\newcommand{\f}{\mathbf{f}}
\newcommand{\m}{\mathbf{m}}
\newcommand{\s}{\mathbf{s}}
\newcommand{\bu}{\mathbf{u}}
\newcommand{\x}{\mathbf{x}}
\newcommand{\y}{\mathbf{y}}
\newcommand{\z}{\mathbf{z}}
\newcommand{\w}{\mathbf{w}}

\newcommand{\radius}{\mathbf{r}}

\newcommand{\cF}{\mathcal F}
\newcommand{\fd}{\mathcal{F}_{d}}
\newcommand{\fz}{\mathcal{F}_{z}}

\newcommand{\OURS}[0]{\textbf{OURS}}
\newcommand{\FGSMU}[1]{\textbf{FGSM-U(#1)}}
\newcommand{\FGSMT}[1]{\textbf{FGSM-T(#1)}}
\newcommand{\FGSMUE}[1]{\textbf{FGSM-UE(#1)}}
\newcommand{\FGSMTE}[1]{\textbf{FGSM-TE(#1)}}

\newcommand\extrafootertext[1]{%
    \bgroup
    \renewcommand\thefootnote{\fnsymbol{footnote}}%
    \renewcommand\thempfootnote{\fnsymbol{mpfootnote}}%
    \footnotetext[0]{#1}%
    \egroup
}

\newcommand{\colvecTwo}[2]{\ensuremath{
		\begin{bmatrix}{#1}	\\	{#2}	\end{bmatrix}
}}
\newcommand{\colvec}[3]{\ensuremath{
		\begin{bmatrix}{#1}	\\	{#2}	\\	{#3} \end{bmatrix}
}}
\newcommand{\colvecFour}[4]{\ensuremath{
		\begin{bmatrix}{#1}	\\	{#2}	\\	{#3} \\	{#4}	\end{bmatrix}
}}

\newcommand{\rowvecTwo}[2]{\ensuremath{
		\begin{bmatrix}{#1}	&	{#2}	\end{bmatrix}
}}
\newcommand{\rowvec}[3]{\ensuremath{
		\begin{bmatrix}{#1} &	{#2}	&	{#3} \end{bmatrix}
}}
\newcommand{\rowvecFour}[4]{\ensuremath{
		\begin{bmatrix}{#1}	&	{#2}	&	{#3} &	{#4}	\end{bmatrix}
}}

\newcommand{\tr}{^\intercal}

\newcommand{\rg}{\mathbf{rk}}
\newcommand{\fA}{f_{\mbs \Theta_A}}
\newcommand{\fB}{f_{\mbs \Theta_B}}
\newcommand{\tA}{\mbs \Theta_A}
\newcommand{\tB}{\mbs \Theta_B}

\newcommand{\PPc}[1]{\textcolor{orange}{[\textbf{Pat}: #1]}}
\newcommand{\PP}[1]{\textcolor{orange}{#1}}
\newcommand{\MCc}[1]{\textcolor{blue}{[\textbf{Matt}: #1]}}
\newcommand{\MC}[1]{\textcolor{blue}{#1}}
\newcommand{\MEc}[1]{\textcolor{green}{[\textbf{Martin}: #1]}}
\newcommand{\LC}[1]{\textcolor{cyan}{#1}}
\newcommand{\mbf}[1]{\ensuremath{\mathbf{#1}}}
\newcommand{\mbs}[1]{\ensuremath{\boldsymbol{#1}}}
\newcommand{\mcl}[1]{\ensuremath{\mathcal{#1}}}
\newcommand{\mrm}[1]{\ensuremath{\mathrm{#1}}}
\newcommand{\mbb}[1]{\ensuremath{\mathbb{#1}}}
\newcommand{\msf}[1]{\ensuremath{\mathsf{#1}}}

\newcommand{\ve}[1]{\ensuremath{\mathbf{#1}}} 

\newcommand{\trsp}[1]{\ensuremath{#1^{\top}}}
\newcommand{\pinv}[1]{\ensuremath{#1^{\dagger}}}
\newcommand{\bmat}[4]{\ensuremath{\begin{bmatrix}#1&#2\\#3&#4\end{bmatrix}}}
\def\trace{\ensuremath{\mathrm{trace}}}
\def\deter{\ensuremath{\mathrm{det}}}
\def\diag{\ensuremath{\mathrm{diag}}}
\def\rank{\ensuremath{\mathrm{rank}}}
\def\Id{\m{Id}} 
\def\mA{\m{A}}
\def\mB{\m{B}}
\def\mC{\m{C}}
\def\mD{\m{D}}
\def\mE{\m{E}}
\def\mF{\m{F}}
\def\mG{\m{G}}
\def\mH{\m{H}}
\def\mK{\m{K}}
\def\mL{\m{L}}
\def\mN{\m{M}}
\def\mP{\m{P}}
\def\mW{\m{W}}
\def\mX{\m{X}}
\def\mY{\m{Y}}
\def\mZ{\m{Z}}

\newcommand{\bvec}[2]{\ensuremath{\begin{bmatrix}#1\\#2\end{bmatrix}}}
\def\One{\mbs{1}} 
\def\va{\ve{a}}
\def\vb{\ve{b}}
\def\vc{\ve{c}}
\def\vd{\ve{d}}
\def\vf{\ve{f}}
\def\vg{\ve{g}}
\def\vh{\ve{h}}
\def\vi{\ve{i}}
\def\vt{\ve{t}}
\def\bx{\ve{x}}
\def\by{\ve{y}}
\def\bz{\ve{z}}
\def\bv{\ve{v}}

\def\cL{\mcl{L}}

\def\ie{\emph{i.e.}}
\def\eg{\emph{e.g.}}
\def\iid{\emph{i.i.d.}}
\def\wrt{w.r.t.}
\def\mwrt{\mrm{w.r.t.}}
\def\msbt{\mrm{sb.t.}}

\def\sqt{^{\frac{1}{2}}} 
\def\msqt{^{-\frac{1}{2}}} 
\def\R{\mbb R}
\def\vtheta{\mbs{\theta}}


\title{Multi-view Tracking Using Weakly Supervised Human Motion Prediction}

\author{Martin Engilberge\\
EPFL, Lausanne, Switzerland \\
{\tt\small martin.engilberge@epfl.ch}
\and
Weizhe Liu\\
Tencent XR Vision Labs \\
{\tt\small weizheliu@tencent.com}
\and
Pascal Fua\\
EPFL, Lausanne, Switzerland \\
{\tt\small pascal.fua@epfl.ch}
}

\maketitle
\thispagestyle{empty}

\begin{abstract}

Multi-view approaches to people-tracking have the potential to better handle occlusions than single-view ones in crowded scenes. They often rely on the \textit{tracking-by-detection} paradigm, which involves detecting people first and then connecting the detections. In this paper, we argue that an even more effective approach is to predict people motion over time and infer people's presence in individual frames from these. This enables to enforce consistency both over time and across views of a single temporal frame. We validate our approach on the PETS2009 and WILDTRACK datasets and demonstrate that it outperforms state-of-the-art methods.

\end{abstract}

\extrafootertext{Project code at \url{https://github.com/cvlab-epfl/MVFlow}}

 \section{Introduction}\label{sec:intro}

\begin{figure}[!ht]
  \centering
    \begin{subfigure}{.4\textwidth}
      \centering
      \includegraphics[width=\linewidth]{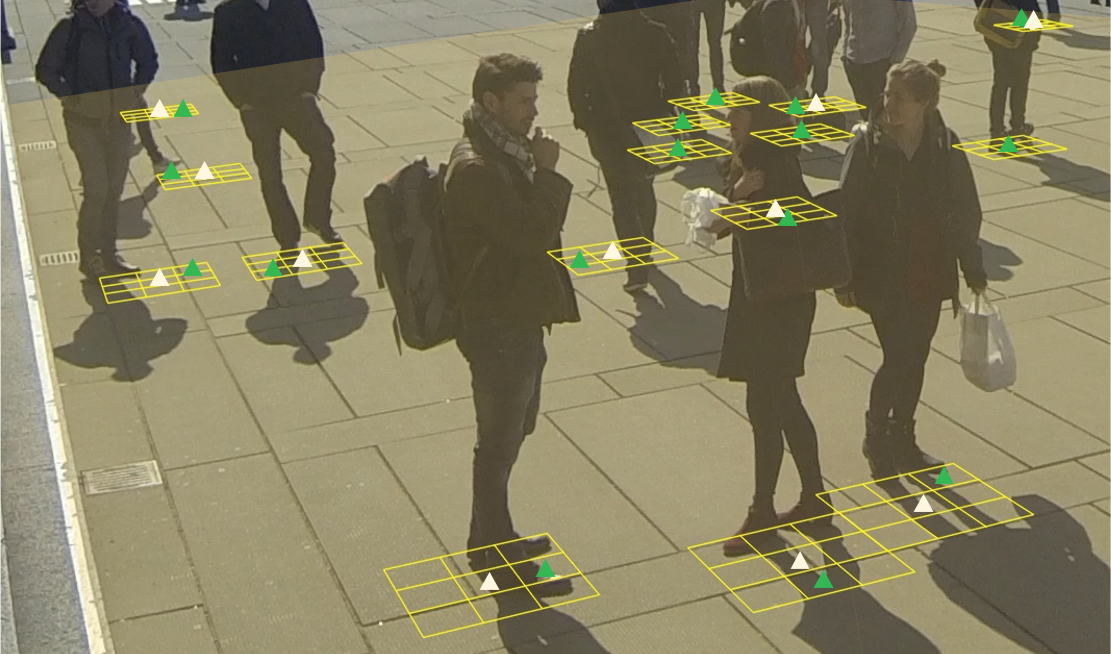}
    \end{subfigure}%
    \vskip 7pt
    \begin{subfigure}{.4\textwidth}
      \centering
      \includegraphics[width=\linewidth]{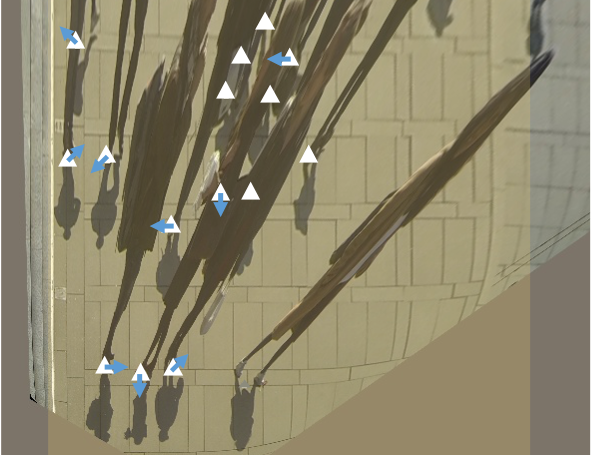}
    \end{subfigure}
  
    \caption{\small \textbf{Predicting human motion.} Our model learns to detect people by predicting human flows. It generates the probabilities that a person moves from one location to one of its eight neighbors or itself, depicted by the yellow grid in the top image. The white triangles depict detections in the ground-plane while the green ones denote the predicted location at the next time step. The bottom image corresponds to the top view re-projection of the  top one, the blue arrows illustrate the motion predicted by our model. On both images the region of interest is overlaid in yellow. People outside of that region are ignored.}
    \label{fig:intro}
    \vspace{-0.5em}
  \end{figure}

When it comes to tracking multiple people,  tracking-by-detection~\cite{Andriluka08}  has  become a  standard paradigm and has  proven effective  for many  applications  such  as surveillance  or  sports player tracking.  It involves first detecting  the target  objects in individual  frames,  associating  these  detections  into  short  but  reliable trajectories  known  as  tracklets,  and  then  concatenating  them  into  longer  trajectories~\cite{Long18,Henschel18,Keuper18,Yoon18,Kim18,Maksai19,Saleh21,Pang21b,Kim21a,Tokmakov21,gan2021self}. The grouping of detections into full trajectories  can also be formulated as the search for multiple min-cost paths on a graph~\cite{Berclaz11,Wang19f}. More recently,  tracking-by-regression~\cite{zhou2020tracking,Xu20b} has been advocated as a potential alternative. It readily enables tracking while being end-to-end differentiable, unlike the detection-based approaches.

However, these single-view tracking techniques can be derailed by occlusions and are bound to fragment tracks when detections are missed. Using multiple cameras is one way to address this problem, especially in locations such as sports arenas where such a setup can be installed once and for all~\cite{BenShitrit14,Xu16,Chavdarova18a}. This can be highly effective but can still fail when occlusions become severe. This is in part because detection algorithms typically operate on single frames and fail to exploit the fact that we have videos that exhibit time consistency. In other words, if someone is detected in one frame, chances are they should be found at a neighboring location in the next frame, as depicted by \cref{fig:intro}. Furthermore, even though people's motion and scale is consistent across views,  that consistency is rarely enforced when fusing results from different views.  

In this paper, we address both these issues by training networks to detect {\it flows of people} across images.  Our model directly leverages temporal consistency using self-supervision across video frames. Furthermore, it can fuse information from different cameras while retaining spatial consistency for human position from different viewpoints. As a result, we outperform state-of-the-art multi-view techniques~\cite{hou2021multiview,hou2020multiview,Chavdarova18a} on challenging datasets.

 \section{Related works}\label{sec:related}
 
Early work on tracking objects in video sequence rely on model evolution technique which focuses on tracking a single object using gating and Kalman filtering~\cite{Mittal03a}. Because of their recursive nature, they are prone to errors such as drift, which are difficult to recover from. Therefore, this method is largely replaced by tracking-by-detection techniques which have proven to be effective in addressing people tracking problems. In this section we first briefly introduce previous work of tracking-by-detection and then discuss previous work in modeling human motions.

\subsection{Tracking-by-Detection.} 
Tracking-by-detection~\cite{Andriluka08} aims to track objects in video sequences by optimizing a global objective function over many frames given frame-wise object detection information. They reply on Conditional Random Fields~\cite{Lafferty01,Yang12a,Milan13}, Belief Propagation~\cite{Yedidia00,Choi12}, Dynamic or Linear Programming~\cite{Bellman57,Segal13}, or Network Flow Programming~\cite{Ahuja93,Dehghan15}.  Some of these algorithms follow the graph formulation with nodes as either all the spatial locations where an object can be present~\cite{Fleuret08a,Berclaz11,BenShitrit14} or only those where a detector has fired~\cite{Jiang07,Tang15,Shu12,Benfold11}.

Among these graph-based approaches, the K-Shortest Paths (KSP) algorithm~\cite{Berclaz11} works on the graph of all potential locations over all time instants, and finds the ground-plane trajectories that yield the overall minimum cost. This optimality is achieved at the cost of multiple strong assumption about human motion, in particular it treats all motion direction as equiprobable. Similar to the KSP algorithm, the Successive Shortest Paths (SSP) approach~\cite{Pirsiavash11} links detections using sequential dynamic programming. ~\cite{Lenz15b} extends this SSP approach with bounded memory and computation which enables tracking in even longer sequences. The memory consumption is further reduced in~\cite{Wang19f} by exploiting the special structures and properties of the graphs formulated in multiple objects tracking problems. More recent work~\cite{Wojke17} proposes to learn a deep association metric on a large-scale person re-identification dataset which enables reliable people tracking in long video sequence. 

Occlusion makes it extremely challenging to achieve reliable object tracking in long sequences. Some algorithms address this by leveraging multiple viewpoints, some approaches first detect people in single images before reprojecting and matching detections into a common reference frame \cite{Xu16,Fleuret08a}. \cite{Baque17b} propose to directly combine view aggregation and prediction with a joint CNN/CRF.
More recently \cite{hou2020multiview} proposed to use spatial transformer networks \cite{Jaderberg15} to project feature representation in the ground plane resulting in an end-to-end trainable multi-view detection model. 
\cite{song2021stacked} proposed to combine multiple views using an approximatiion of a 3D world coordinate system by projecting features in planes at different height levels.
Finally \cite{hou2021multiview} proposed to use multi-view data augmentation combined with a transformer architecture to fuse ground plane features from multiple points of view and  obtains state-of-the-art results for multiple object detection  on the WILDTRACK dataset \cite{Chavdarova18a}.

\subsection{Modeling human motion}
Modeling human motion as flow when tracking people has been a concern long before the advent of deep learning~\cite{Pellegrini09,Vogel11,Butt13,Liu13,Collins12,Laptev07b,Gijsberts14,Butt12,Nater11,Milan14,Andriyenko10,Berclaz11}. For example, in~\cite{Berclaz11}, people tracking is formulated as  multi-target tracking on a grid and gives rise to a linear program that can be solved efficiently using the K-Shortest Path algorithm~\cite{Suurballe74}. The key to this formulation is to optimize the people flows from one grid location to another, instead of the actual number of people in each grid location. In~\cite{Pirsiavash11}, a people conservation constraint is enforced and the global solution is found by a greedy algorithm that sequentially instantiates tracks using shortest path computations on a flow network~\cite{Zhang08a}. Such people conservation constraints have since been combined with additional ones to further boost performance. They include appearance constraints~\cite{BenShitrit11,Dickle13,BenShitrit14} to prevent identity switches,  spatiotemporal constraints to force the trajectories of different objects to be disjoint~\cite{He16b}, and higher-order constraints~\cite{Butt13,Collins12}. 
More recent work extends this flow formulation with deep learning~\cite{Liu20a,Liu21a} to formulate people as people flows which contributes to reliable people counting in even dense regions.  However, none of these methods leverage such people flow formulation to address tracking problems with deep neural networks. These kinds of flow constraints have therefore never been used in a deep people tracking context. 
 
 \section{Approach}\label{sec:approach}
 Most recent approaches rely on the \textit{tracking-by-detection} paradigm. 
In its simplest form, the detection step is disconnected from the association step.
In this section, we propose a novel method to bring closer those two steps. First we introduce a detection network predicting people flow in a weakly supervised manner. 
Then we show how we modify existing association algorithms to leverage predicted flows to generate unambiguous tracks.

\subsection{Formalism}
\label{sec:formalism}

Let us consider a multi-view video sequence $S=\{\mathbf{I}^1, \mathbf{I}^2,  ..., \mathbf{I}^{T-1}, \mathbf{I}^T\}$ consisting of $T$ time steps.
Each time step $\mathbf{I}^t=\left\{\mathbf{I}_1^{t}, \ldots \mathbf{I}_V^{t}\right\}$ consists of a set of synchronized frames taken by $V$ cameras with overlapping fields of view.
For each camera the calibration $\mathbf{C}_{v}$ is known and contains both intrinsic and extrinsic parameters.
Each frame $\mathbf{I}_v^t \in(0,255)^{W \times H \times 3}$  is a color image with a spatial size ($W$,$H$).

To combine multiple views we choose to work in the common ground plane. For each frame we define $\mathbf{G}_v^t = P(\mathbf{I}_v^t, \mathbf{C}_{v})$ as the  projection of frame $\mathbf{I}_v^t$ 
on the ground plane using the projection function $P$ producing $\mathbf{G}_v^t \in(0,255)^{w \times h \times 3}$ with ($w$, $h$)  the spatial size of the ground plane image. 

Finally, we adopt similar grid world formalism as previous work \cite{Berclaz11}. At each time step $t$ we discretize the physical ground plane to form a grid of $w \times h$ cells, 
giving us a scene representation of dimensionality ${w \times h \times t}$ for a full sequence.

\subsection{People Flow}
\label{sec:peopleflow}

\begin{figure}
    \centering
      \includegraphics[width=.5\linewidth]{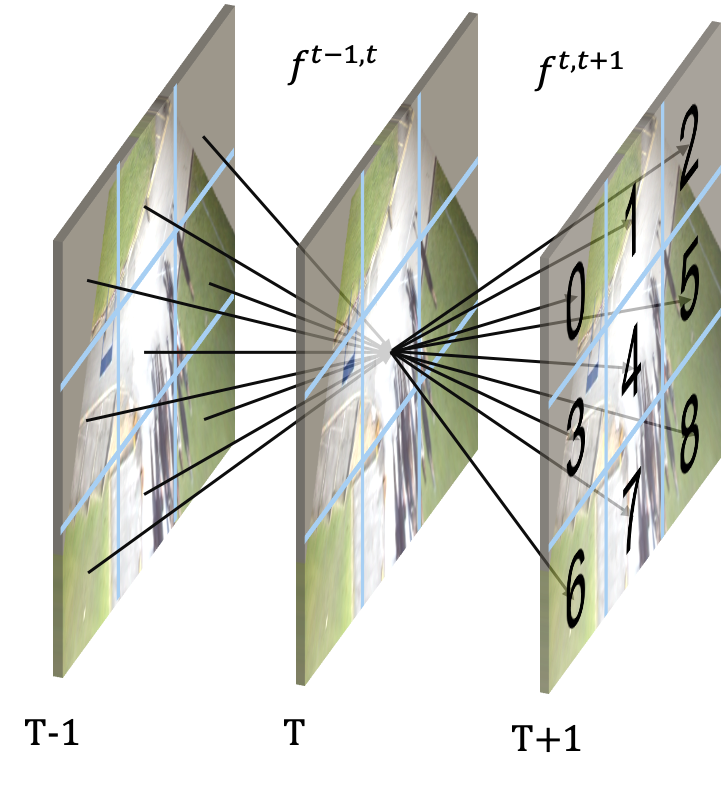}
      \caption{\textbf{ Grid flow representation} For each location $i$ we predict the probability that a person is moving from $i$ to one of its eight neighbors or itself in the next time step.
      Detection probability at a given location at time t can be computed by summing the nine outgoing flows or the nine flows reaching that location from t-1.
      \label{fig:gridworld}
      }
    \end{figure}

Given a pair of consecutive multi-view time steps we define the human flow $f^{t,t+1}$ as follows: For a given location $i$, the flow $f^{t,t+1}_{i,j}$ 
is the probability that a person in cell $i$ at time $t$ moves to location $j$ at time $t+1$. Where $j \in \mathcal{N}(i)$ is a neighbor of $i$. 
Concretely, for each cell in the ground plane we represent people flow by a 9-dimensional vector of probability (one dimension per neighbors of that cell). 
The grid representation and the definition  of neighborhood are illustrated in \cref{fig:gridworld}


To accurately model human motion, the flow need to respect three constraints:

First, people conservation constraints, if a person is present at time $t$, he should be present at time $t+1$ in the same location, 
or in a neighboring one. In other words, if we consider three time steps $\mathbf{I}^{t-1}$, $\mathbf{I}^t$, and $\mathbf{I}^{t+1}$ the sum of the incoming flow in cell $j$ between time $t-1$ and $t$ 
should be equal to the sum of outgoing flow between time $t$ and $t+1$. More formally it reads:
\begin{equation}
    \sum_{i \in \mathcal{N}(j)} f_{i, j}^{t-1, t}=x_{j}^{t}=\sum_{k \in \mathcal{N}(j)} f_{j, k}^{t, t+1}.
\label{eq:conservation}
\end{equation}
The sums of the flow are equal to $x_{j}^{t}$ the probability that there is a person in $j$ at time $t$.

Second, non-overlapping constraints, at any time there should be at most one person in every cell.
\begin{equation}
\forall k, t, \sum_{j \in \mathcal{N}(k)} f_{k, j}^{t,t+1} \leq 1.
\label{eq:overlap}
\end{equation}
Finally, a temporal consistency constraint, if we reversed a sequence, the flow should be the same with the flow direction being flipped.

\begin{equation}
    f_{i, j}^{t-1, t}=f_{j, i}^{t, t-1}.
    \label{eq:temporalconsistency}
\end{equation}
Reconstructing detection from human flow is trivial (\cref{eq:conservation}) and has a unique solution, on the other hand, generating flow from detection can have multiple solutions. Therefore we introduce Multi-View FlowNet (MVFlow), trained to generate human flow.
By predicting flow instead of detection our model is able to take advantage of the asymmetric mapping between flow and detection. It learns to predict flow in a weakly supervised manner, using only detection annotation. 
Enforcing flow constraints in \cref{eq:conservation}, \cref{eq:overlap} and \cref{eq:temporalconsistency} also serves as a regularization for the final detection. Predictions are temporally consistent and represent natural human motion.


\begin{figure*}[t]
\begin{center}
\centering\includegraphics[width=1\textwidth]{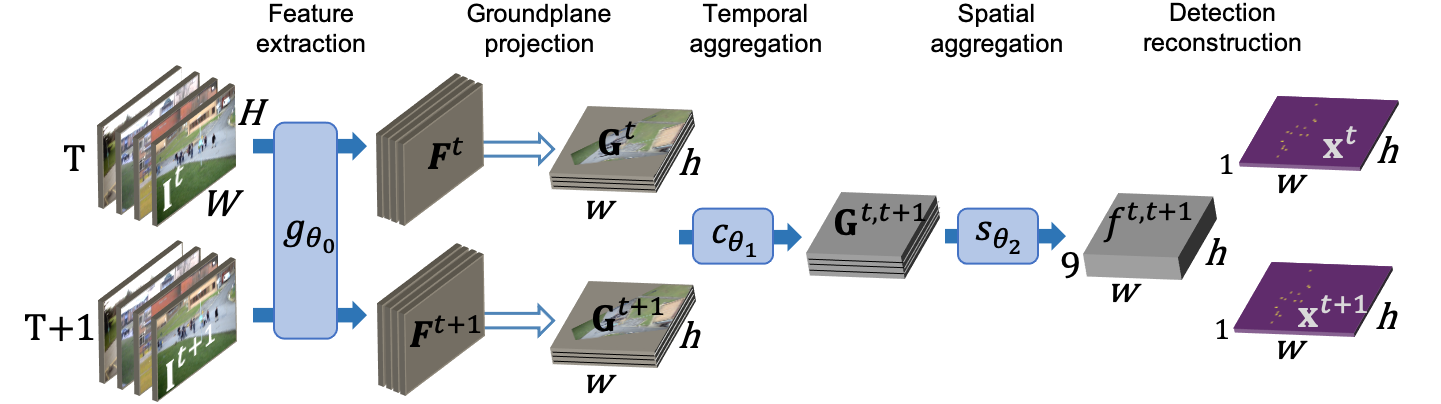}
\end{center}
\caption{\textbf{Proposed Multiview Detection Architecture}. Two consecutive sets of multi-view frames are transformed into human flow $f^{t,t+1}$ 
by the proposed multi-view prediction model.
The human flow is then used to reconstruct detection heatmaps $\ve x^t$ and $\ve x^{t+1}$.
The architecture with parameters $(\mbs \theta_0, \mbs \theta_1, \mbs \theta_2)$ is trained with ground truth detection only.
Blue background boxes are trainable modules (with parameters indicated on top). 
Output dimensions $h=w=128$ is used in the experiments.
}
\label{fig:model}
\end{figure*}

\subsection{Multi-View architecture}

In this section we detail the architecture of MVFlow, our multi-view detection model.

The proposed model consists of 5 steps and takes as input a pair of multi-view frames. Each frame is processed by a ResNet. The resulting features are projected in the ground plane. Ground features from the same point of view at time $t$ and $t+1$ are aggregated. Afterwards, the spatial aggregation module combines the features from the different points of view into human flow.
Detection predictions are reconstructed from the flow for both time steps. The 5 steps are illustrated in \cref{fig:model}.
More formally, the model is defined as follows:

\begin{equation}
\genfrac{}{}{0pt}{}{\ve I^t}{\ve I^{t+1}} \hspace{-0.1cm} \xmapsto{g_{\mbs\theta_0}} \hspace{-0.1cm}
\genfrac{}{}{0pt}{}{\ve F^t}{\ve F^{t+1}} \hspace{-0.1cm} \xmapsto{\mathbf{C}} \hspace{-0.1cm}
\genfrac{}{}{0pt}{}{\ve G^t}{\ve G^{t+1}} \hspace{-0.1cm} \xmapsto{c_{\mbs\theta_1}} \hspace{-0.1cm}
\ve G^{t,t+1} \hspace{-0.1cm} \xmapsto{\mathrm{s_{\mbs\theta_2}}} \hspace{-0.1cm}
f^{t,t+1}  \hspace{-0.1cm} \xmapsto{\mathrm{rec.}} \hspace{-0.1cm}
\genfrac{}{}{0pt}{}{\ve x^t}{\ve x^{t+1}},
\label{eq:pipe}
\end{equation}
Where:
$g_{\boldsymbol{\theta}_{0}}(\ve I^t_v) \in \mathbb{R}^{W/8 \times H/8 \times D}$ is the output of the ResNet parametrized by weights $\boldsymbol{\theta}_{0}$. $P(\mathbf{F}_v^t, \mathbf{C}_{v})$ project features onto the groundplane.
Temporal aggregation is achieved with a convolution parametrized by weights $\boldsymbol{\theta}_{1}$ and output $c_{\boldsymbol{\theta}_{1}}(\ve G^t_v, \ve G^{t+1}_v) \in \mathbb{R}^{w \times h \times D'}$. 
The spatial aggregation layer parametrized by weights $\boldsymbol{\theta}_{2}$ generates the human flow $s_{\boldsymbol{\theta}_{2}}(\ve G^{t,t+1}) \in \mathbb{R}^{w \times h \times 9}$ .
The detection heatmaps $\ve x \in \mathbb{R}^{w \times h} $ are reconstructed from the flow as follows,

\begin{equation}
    \begin{split}
    x_{j}^{t}=\sum_{k \in \mathcal{N}(j)} f_{j, k}^{t, t+1},  \hspace{5mm}  \bar x_{j}^{t}=\sum_{k \in \mathcal{N}(j)} f_{k, j}^{t+1, t}.
    \end{split}
    \label{eq:detreconstruction}
\end{equation}
We denote by $\bar x_{j}^{t}$ the reconstruction obtained by reversed flow. According to \cref{eq:temporalconsistency}, 
for every forward flow there is an equivalent reversed flow. The inverted flow $f^{t+1, t}$ is obtained by swapping the inputs of the model.


\paragraph{Spatial Aggregation} The use of ground plane representations greatly simplifies the spatial aggregation of multiple views. 
However the camera calibrations are never perfect and minor misalignments between views are common. Our aggregation mechanism is designed to be robust to such misalignment.
The feature representations for all the points of view are concatenated alongside their channel dimension, then a convolutional layer with large kernel size ($5 \times 5$) allows for realignment between neighboring features. 
An efficient spatial aggregation mechanism also needs to handle occlusion, objects hidden in some views should still be predicted from the rest of the views.
To make this process easier we propose a multi-scale module: occlusions are easier to detect at a coarse level, while fine level features are needed for precise localization.
The multi-scale feature works as follows: The features are pooled to 4 different sizes, and are processed by  four sets of convolutions, batch normalizations and ReLUs, one for each scale.
The 4 scale representations are then upscaled back to the same dimension and combined with a convolutional layer. 
Finally, a convolutional layer followed by a sigmoid activation function transform the aggregated feature into the human flow $f^{t-1, t}$.

\subsection{Training}

The goal of the training is to learn the parameters $\boldsymbol{\theta}_{0: 2}$. Through a combination of loss functions, we aim at applying the constraints defined in \cref{sec:peopleflow}.
Each frame $\mathbf{I}_V^{t}$ is annotated with the position of every human foot. Using $P$ we obtain the 2d coordinate of every human in the ground plane
 and generate binary ground truth detection maps $\ve y^t \in (0,1)^{w \times h}$. Note that the ground truth heatmaps are independent from the point of view.

Given two multiview set of frames $\mathbf{I}^t$, $\mathbf{I}^{t+1}$ and their respective ground truth detection map $\ve y^t$, $\ve y^{t+1}$ we define the loss function as follows.
\begin{align}
    L =  L_{\text {det }}(\ve x^t, \ve y^t) +  L_{\text {det }}(\ve x^{t+1}, \ve y^{t+1}) \nonumber \\
  + L_{\text {cycle }}(f^{t,t+1},f^{t+1,t}),
    \label{eq:loss}
\end{align}
with $ L_{\text {det }}$ as the detection loss between prediction from flow and inverted flow and ground truth detection.
Applied both at time $t$ and $t+1$ together $ L_{\text {det }}$ enforces the constraints defined in \cref{eq:conservation} and \cref{eq:overlap}.
\begin{equation}
    L_{\text {det }}(\ve x^t, \ve y^t) = \left(\ve x^{t}-\ve y^t\right)^{2} + \left(\ve {\bar x}^{t}- \ve y^t\right)^{2}.
    \label{eq:lossdet}
\end{equation}
The temporal consistency constraint defined in \cref{eq:temporalconsistency} is directly applied on the flow with the loss $L_{\text {cycle }}$ between the flow and its inverted counterpart.
\begin{equation}
    L_{\text {cycle }}(f^{t,t+1},f^{t+1,t}) = \sum_{j \in G^{t}}\sum_{k \in \mathcal{N}(j)}\left(f_{j, k}^{t, t+1}-f_{k, j}^{t+1, t}\right)^{2},
    \label{eq:losscycle}
\end{equation}
where $j \in G$ correspond to every cell in ground plane grid.

\paragraph{Dealing with unbalanced flow} When dealing with video sequences with large frame rates, people barely move between consecutive frames.
It results in a large imbalance in terms of flow, with static flow being occasionally 20 times more common than any other flow direction.
To help with the issue we introduce a motion-based reweighting of our detection loss which reads as follows:
\begin{equation}
L_{\text {det }}(\ve x^t, \ve y^t) = (\left(\ve x^{t}-\ve y^t\right)^{2}  + \left(\ve {\bar x}^{t}- \ve y^t\right)^{2} )\times (1 + \lambda_r|\ve y^t - \ve y^{t+1}|),
\label{eq:reweighting}
\end{equation}
where $|y^t - y^{t+1}|$ correspond to the ground truth detection map containing only the people moving between time $t$ and $t+1$. 
$\lambda_r$ controls the strength of the regularization. Larger values will penalize strongly incorrect prediction of people in motion.
With $\lambda_r=0$ no regularization is applied and we get the original detection loss defined in \cref{eq:lossdet}.

\subsection{Track reconstruction}
\label{sec:kspflow}
We have introduced a detection model that is able to produce detection heatmaps as well as human flow.
In this part we extend two existing association algorithm to leverage human flow while generating tracks.

\paragraph*{}
In KSP\cite{Berclaz11} they reformulate the association problem as a constrained flow optimization problem.
Starting from detection probability maps, they build a dense graph containing all possible location across time.
Each location is connected to its neighbors in the previous and next time step. 
The weight on the edges is proportional to the detection probability at this location.
All the locations in the first time step are connected to a source node, and all the locations in the last time step are connected to a sink node.

The optimization on this graph is maximizing the number of paths connecting the source to the target while minimizing the overall cost.
Given the graph previously defined this can be done using Suurballe's algorithm \cite{Suurballe74}.

One limitation of KSP, is that when building the graph it assumes an equal probability for all the different directions (neighbors).
In crowded scene such an assumption can easily result in identity switches.
We propose to reformulate KSP to have separate probabilities for each direction. We replace Eq. 12 from the original paper \cite{Berclaz11}, 
in order to integrate the predicted flow from our model.
The edge cost for the graph is derived from our predicted flow as follows:

\begin{equation}
    c_{KSPFlow}\left(e_{i, j}^{t}\right)=-\log \left(\frac{f_{i, j}^{t, t+1}}{1-f_{i, j}^{t, t+1}}\right).
    \label{eq:edge_cost_ksp}
\end{equation}
Everything else is kept as in the original KSP paper. We will denote this new method KSPFlow for the rest of the paper.

\paragraph*{}
MuSSP\cite{Wang19f} propose to solve the association step using a min-cost flow method.
As opposed to KSP, a sparse graph is built from pre-extracted detection instead of the full-probability map. It improves computation efficiency 
and allow to model longer range of motion. However the edge costs of the graph only take into consideration, the spatial and temporal distance between detection.
We propose to update the cost formulation to use the flow predicted by our model as follows:
\begin{multline}
    c_{muSSPFlow}\left(e_{i, j}^{t}\right)=-e^{-(\delta_t-1)*\sigma_t}*e^{-d(\ve{x_i},\ve{x_j})*\sigma_d}\\
    *e^{-d(\ve{x_i}+\delta_t*f_{i, j}^{t, t+1},\ve{x_j})*\sigma_f},
    \label{eq:edge_cost_mussp}
\end{multline}
where $\delta_t$ is the temporal distance between detection $i$ and detection $j$ and $d(\ve{x_i},\ve{x_j})$ is the physical distance between them.
We propose to add the term $d(\ve{x_i}+\delta_t*f_{i, j}^{t, t+1},\ve{x_j})$ which model the distance between detection $j$ and the estimated position of detection 
$i$ in the future based on the flow prediction. $\sigma_t$, $\sigma_d$ and $\sigma_f$ are the hyperparameters that respectively control the contribution of the temporal, 
spatial and motion-based distances between the detections.

 \section{Experiments}\label{sec:expe}

To validate our approach, we compare it to state-of-the-art ones on the multi-view multi-object detection and tracking using PETS2009 and WILDTRACK datasets.



\begin{figure*}[h]

    \centering
    \begin{subfigure}[c]{0.235\textwidth}
        \centering
        \includegraphics[width=\linewidth]{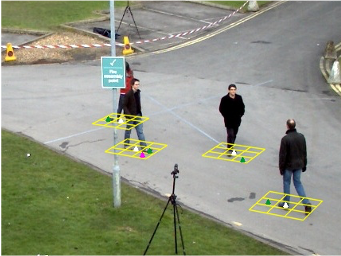}
    \end{subfigure}
    \hfill
    \begin{subfigure}[c]{0.235\textwidth}  
        \centering 
        \includegraphics[width=\linewidth]{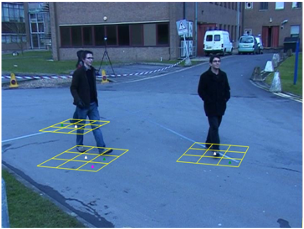}
    \end{subfigure}
    \hfill
    \begin{subfigure}[c]{0.235\textwidth}   
        \centering 
        \includegraphics[width=\linewidth]{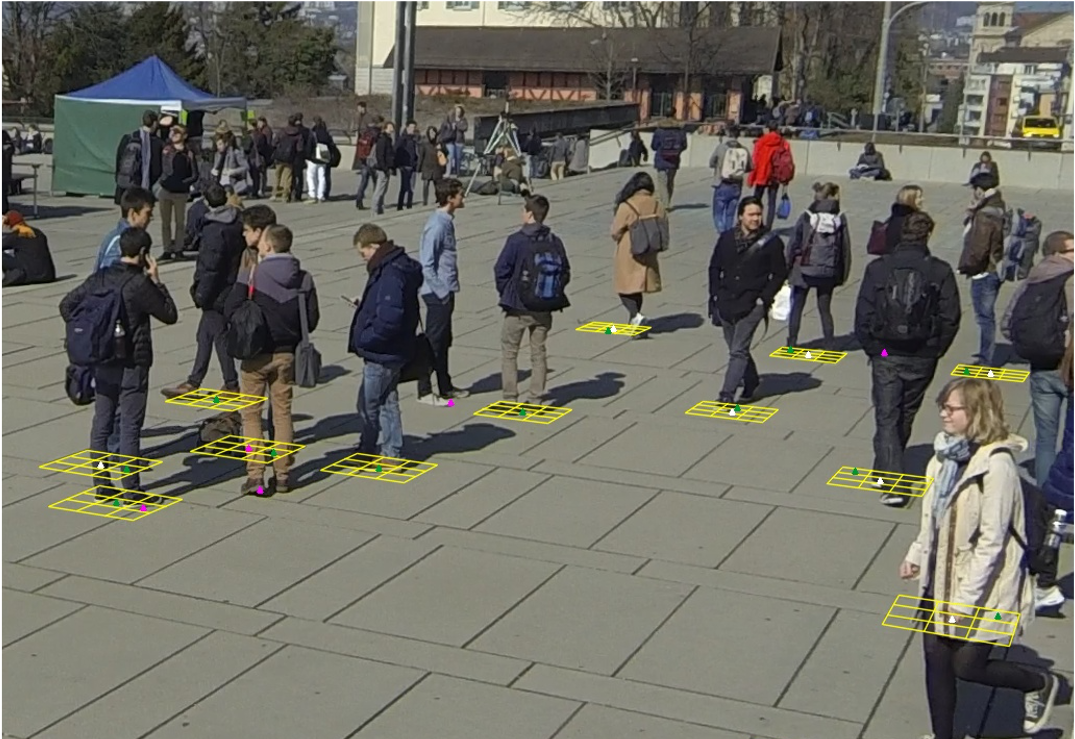}
    \end{subfigure}
    \hfill
    \begin{subfigure}[c]{0.235\textwidth}   
        \centering 
        \includegraphics[width=\linewidth]{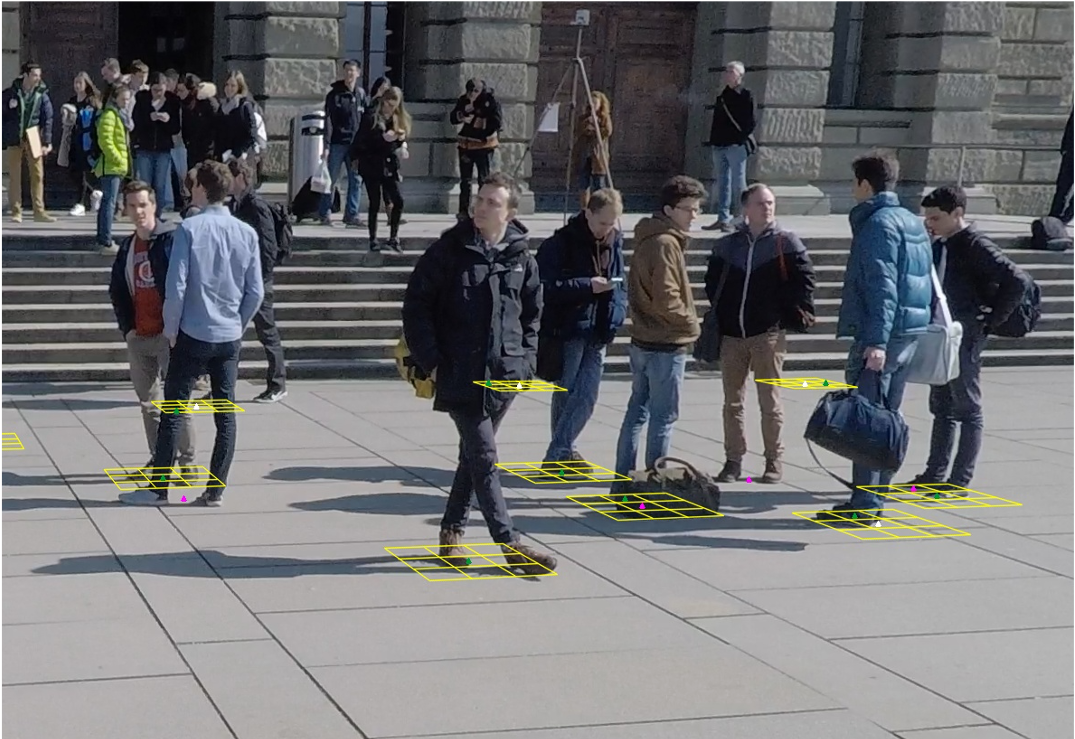}
    \end{subfigure}

\caption{\textbf{Visualization of the predicted flow, viewed best zoomed in.}. For each detected person in the image, we visualize the predicted flow.
Centered around each detection we reproject a $3 \times 3$ grid corresponding to the ground plane division defined in \cref{sec:peopleflow}.
The green triangle mark the cell of the flow direction with the highest probability. Or, in other words, the predicted position for the next time step.
If the prediction is incorrect, a pink dot marks the true destination. Note that ground truth flow is used for visualization purpose only and is never used during training.
The two left images are different points of view from the PETS2009 dataset, the two right images are coming from the WILDTRACK dataset.
}
\label{fig:visu_flow}

\end{figure*}

\subsection{Datasets, Metrics, and Training}

We endeavored to use the most recent and up-to-date baseline and dataset to evaluate our multi-view approach. However, we were limited by the availability of the kind of multi-view datasets we need.

\begin{table}[h]
    \begin{center}
    \begin{tabular}{r  c c c   c c c } \toprule

    &  \multicolumn{6}{c}{WILDTRACK dataset}  \\ \cmidrule(lr{.75em}){2-7}
    model & MODA & Prec. & Rec.  \\ \midrule
    DeepOcclusion \cite{Chavdarova18a} & 74.1 & 95.0 & 80.0 \\
    MVDet \cite{hou2020multiview} & 88.2 & 94.7 & 93.6   \\
    SHOT \cite{song2021stacked} & 90.2 & 96.1 & 94.0   \\
    MVDeTr \cite{hou2021multiview} & 91.5 & \textbf{97.4} & 94.0   \\
    MVFlow (Ours) & \textbf{91.9} & 96.4 & \textbf{95.7}  \\
    \bottomrule
    \end{tabular}
    \end{center}
      \caption{\textbf{Multi-view multi-person detection}  Detection performance of our proposed MVFlow model on the WILDTRACK dataset. MODA, precision and recall are reported.
      }  
     \label{tab:mvmdetection}
    \end{table}

\paragraph{Datasets.}

To train our model we used two datasets of calibrated multiview video sequences. The WILDTRACK dataset \cite{Chavdarova18a} consists of 400 annotated frames with 7 points of view covering a crowded square. As in~\cite{Chavdarova18a}, we use the first 360 frames for training and validation and the last 40 for testing. From the PETS2009 dataset~\cite{Ferryman09a}, we use the sequence S2L1, which consists of 795 frames from 7 points of view. It has a low density and consists of 10 people walking on a road in different directions. We use the first half of the frames for training and test on the second half. This is a departure from the standard evaluation procedure that uses S2L1 only for evaluation. We did this because our algorithm needs to be trained on the same set of views it is tested. Note that our approach is trained from scratch using only the dataset mentioned above while other works commonly rely on pretrained object detectors.


\paragraph{Metrics.}

We rely on the standard CLEAR MOT metric~\cite{Kasturi09}. We report Multiple Object Detection Accuracy (MODA), Multiple Object Tracking Accuracy (MOTA), and Multiple Object Tracking Precision (MOTP). We also include identity preservation metric IDF1~\cite{Ristani16}, computed similarly to the F1 score using the Identity Precision (IDP) and Identity Recall (IDR). To compute these metrics we use the publicly available \textit{py-motmetrics} implementation.

\paragraph{Baselines.}

For detection purposes, we compare to DeepOcclusion\cite{Chavdarova18a}, MVDet\cite{hou2020multiview}, SHOT\cite{song2021stacked} and MVDeTr\cite{hou2021multiview} which are four approaches that tackle multi view detection on the WILDTRACK dataset. As in these papers, the detection are evaluated directly in the groundplane.

For linking purposes, we use DeepSort \cite{Wojke17}, muSSP \cite{Wang19f} and KSP \cite{Berclaz11} for which code is publicly available. Since we are only interested in linking pointwise detections, we ignore appearance models and bounding boxes during tracking. For computational reasons we generate the tracks in two steps, first we generate tracklets up to a length of 40 detections. Then we merge the tracklets to form the final tracks.
We provide additional tracking baseline using DeepSort, muSSP and KSP on the detection results of MVDet and MVDeTr.

\begin{table*}[h]
    \begin{center}
    \begin{tabular}{r  c c c c c c c } \toprule
        &  \multicolumn{6}{c}{PETS S2L1 dataset}  \\ \cmidrule(lr{.75em}){2-7}
    model & MOTA & MOTP & IDF1 & IDP & IDR & ML & MT   \\ \midrule
    B\&P \cite{Leal-Taixe12} & 76 & - & - & - & - & - & -  \\ 
    POM + KSP \cite{Berclaz11} & 78 & - & - & - & - & - & -  \\ 
    HTC \cite{Xu16} & 89 & -  & - & - & - & - & -  \\ 
    MVFlow + deepSORT (Ours) & 71.8 & 1.05 & 72.0 & 64.3 & 81.7 & 0 & 8 \\
    MVFlow + KSP (Ours) & 88.8 & 0.67 & 83.3 & 82.6 & 84.0 & 0 & 7  \\
    MVFlow + KSPFlow (Ours) & 91.4 & 0.67 & \textbf{88.3} & \textbf{90.7} & \textbf{85.9} & 0 & 7  \\
    MVFlow + muSSP (Ours) & 92.2 & 0.65 & 78.5 & 79.6 & 77.4 & 0 & 8  \\
    MVFlow + muSSPFlow (Ours) & \textbf{93.3} & \textbf{0.64} & 84.0 & 85.1 & 82.8 & 0 & 8  \\
    \vspace{-0.5em}
    \rule{0pt}{4ex}  \\
    &  \multicolumn{6}{c}{WILDTRACK dataset}  \\ \cmidrule(lr{.75em}){2-7}
    model & MOTA & MOTP & IDF1 & IDP & IDR & ML & MT   \\ \midrule
    DeepOcclusion+KSP \cite{Chavdarova18a} & 69.6 & - & 73.2 & 83.8 & 65.0 & - & -   \\ 
    DeepOcclusion+KSP+ptrack \cite{Chavdarova18a} & 72.2 & - & 78.4 & 84.4 & 73.1 & - & -  \\ 
    \midrule
    MVDet \cite{hou2020multiview} + deepSORT & 8.3 & 0.95 & 48.8 & 46.9 & 50.8 & 20 & 12 \\
    MVDet \cite{hou2020multiview} + KSP & 46.9 & 0.81 & 64.8 & 95.5 & 49.0 & 28 & 13 \\
    MVDet \cite{hou2020multiview} + muSSP & 80.6 & 0.80 & 79.4 & 79.2 & 79.6 & 4 & 29 \\ 
    \midrule
    MVDeTr \cite{hou2021multiview} + deepSORT & 24.0 & 0.94 & 56.3 & 56.4 & 56.2 & 20 & 14 \\
    MVDeTr \cite{hou2021multiview} + KSP & 48.5 & 0.64 & 65.8 & 97.0 & 49.8 & 28 & 13 \\
    MVDeTr \cite{hou2021multiview} + muSSP & 89.4 & 0.58 & 90.7 & 90.5 & 90.9 & 3 & 33 \\  
    \midrule
    MVFlow + deepSORT (Ours) & 52.8 & 0.89 & 73.3 & 65.2 & 83.7 & \textbf{2} & 33  \\
    MVFlow + KSP (Ours) & 81.9 & 0.61 & 79.8 & 80.2 & 79.5 & 4 & 32  \\
    MVFlow + KSPFlow (Ours) & 83.5 & 0.61 & 81.0 & 81.7 & 80.3 & 5 & 29  \\
    MVFlow + muSSP (Ours) & \textbf{91.3} & \textbf{0.57} & \textbf{93.5} & \textbf{92.7} & \textbf{94.2} & \textbf{2} & \textbf{38} \\
    MVFlow + muSSPFlow (Ours) & 91.2 & \textbf{0.57} & 93.4 & 92.6 & \textbf{94.2} & \textbf{2} & \textbf{38} \\
   

    \bottomrule
    \end{tabular}
    \end{center}
      \caption{\textbf{Multi-view multi-person tracking} Tracking performance of our proposed MVFlow model
      on two datasets: PETS2009 S2L1 and WILDTRACK. For each dataset we show the result of our detection model combined with 
      five different association algorithms. Our model MVFlow + muSSPFlow achieves state-of-the-art results both on PETS2009 and WILDTRACK.
      We report the CLEAR MOT metrics MOTA, MOTP, Mostly Loss (ML), Mostly Track (MT), as well as identity conservation metrics IDF1, IDP and IDR.
      }  
     \label{tab:mvmtracking}
    \end{table*}


\paragraph{Model Training.}

Our model trained using the Adam optimizer \cite{Kingma15} with a batch size of one and a learning rate of $0.001$. The learning rate is halved after epoch 20, 40, 60, 80 and 100. 
Random rectangular crops are taken from the input image and resized to a fixed-size of  $536 \times 960$. The ResNet 34 \cite{He16a} reduces this dimensionality by a factor 8.
The ground plane projection $P$ produce the final spatial dimension of the flow and detection output of $w=128, h=128$.

\subsection{Comparing to the State-of-the-Art}

\paragraph{Multiview Detection.}

Our model builds detection heatmaps from human flow. To produce actual detections from these heatmaps, we use Non Maximum Suppression (NMS). We then select the top-200 and use K-mean clustering to filter-out spurious ones.

We report our results on the WILDTRACK dataset in \cref{tab:mvmdetection}. Our model consistently outperforms the baselines.  In particular it improves MODA by 0.4 over MVDeTr~\cite{hou2021multiview}, the previously best performing model that uses a complex transformer architecture and additional supervision in the image plane.

\paragraph{Multiview Tracking.} 

We report our tracking results on both WILDTRACK and PETS 2009 in \cref{tab:mvmtracking}. We report result for 5 associations algorithm, 3 of which are baselines: DeepSort is online, while KSP and muSSP are offline and graph based. KSPFlow and muSSPFlow introduced in \cref{sec:kspflow} are implemented by replacing the edge cost function of KSP and muSSP with \cref{eq:edge_cost_ksp} and \cref{eq:edge_cost_mussp}. 

Combining MVFlow with muSSPFlow delivers the best MOTA. Note that both KSPFlow and muSSPFlow respectively match or outperform KSP and muSSP both in terms of MOTA and IDF1 on both datasets.
This gap in performance clearly confirm the benefit of predicting human motion and using it for tracking.



\subsection{Ablation Study}
To evaluate the contribution of the different component of the proposed approach we conduct and ablation study. We use the WILDTRACK dataset and combine the ablated model with muSSP to generate tracking results.  We report the results in \cref{tab:ablation} and the performance numbers should be compared to the WILDTRACK numbers in the penultimate row of \cref{tab:mvmtracking}.
\begin{table*}[h]
    \begin{center}
    \begin{tabular}{r  c c c c c c c c } \toprule
    
    model & MOTA & MOTP & IDF1 & IDP & IDR & ML & MT &  MODA \\ \midrule
    $\lambda_r = 0$ & 58.5 & 0.62 & 62.0 & 73.8 & 53.4 & 18 & 13 & 61.1 \\
    No aug. & 76.8 & 0.65 & 73.0 & 76.0 & 70.2 & 4  & 29 & 80.6 \\
    Single scale & 85.3 & 0.59 & 85.5 & 87.1 & 84.0 & 5  & 33 & 87.9 \\
    No flow & 87.4 & 0.61 & 90.6 & 88.4 & 92.8 & 3 & 40 & 88.6  \\
    No $L_{\text {cycle }}$ & 87.6 & 0.61 & 86.7 & 85.7 & 87.8 & 5 & 38 & 89.3  \\
    $\lambda_r = 10$ & 87.8 & 0.59 & 86.4 & 86.9 & 85.9 & 5 & 35 & 89.7  \\
    \bottomrule
    \end{tabular}
    \end{center}
      \caption{\textbf{Ablation result on multi-view multi-person tracking} Using the WILDTRACK dataset, 
      we test how each component of our model contributes to the overall performance. We remove: the flow balancing term in the loss ($\lambda_r = 0$), the data augmentation, the multi-scale view aggregation and replace it with a single scale one. We replace the flow by directly regressing detection, we remove the temporal consistency term of our loss. All the results are computed using the muSSP association algorithm (comparable to the penultimate row of \cref{tab:mvmtracking})
      }  
     \label{tab:ablation}
     \vspace{-0.5em}
    \end{table*}

\paragraph*{No Flow Prediction.}
To ignore the flow, we modify the last convolutional layer of our model such that it directly produces detection heatmaps.  In other words, we remove the flow prediction and detection reconstruction steps, everything else is kept as is. As can be seen by comparing the fourth row of \cref{tab:ablation} to the penultimate row of \cref{tab:mvmtracking}, the performance decreases substantially, which confirms the importance of flow prediction. This experiment demonstrates the regularization property of our flow formulation. Predicting the flow under realistic motion constraints improve the reconstructed detection on its own, even without explicitly using the flow.
\paragraph*{No Temporal Consistency}
Temporal consistency as defined in \cref{eq:temporalconsistency} enforces similarity between flow and its inverted counterpart. It forces the network to model temporal information about the order of frames and to not purely rely on visual cues. As can be seen in \cref{tab:ablation}, removing it during training reduce both MOTA and identity preservation measured by IDF1. 

\paragraph*{Flow Reweighting}

Across the different datasets, we observed a dominance of static flow over others. To avoid overfitting to the most common flow direction, we introduced in \cref{eq:reweighting} a motion-based reweighting scheme in our loss function. We now test different values for $\lambda_r$ that controls how much  re-weighting there is. Our best model was obtained with $\lambda_r=5$. In the first and last rows of \cref{tab:ablation}, we report results with  $\lambda_r=0$---no re-reweigthing---and $\lambda_r=10$. The performance is degraded in both cases, especially when  $\lambda_r=0$. Note that the training remains robust over a wide range of $\lambda_r$ and value between one and six all gave decent results.

\paragraph*{Single Scale Spatial Aggregation.}

Robust view aggregation requires the ability to handle misalignment between views. To do so we proposed a multiscale aggregation mechanism. To test its efficiency we only retain the highest resolution branch in our aggregation module. Everything else is kept the same. As shown in the third row of \cref{tab:ablation}, this also degrades performance, thus confirming the importance  multiscale aggregation. Doing the aggregation at multiple scales helps to deal with slight misalignments between views, the model learns which scale contains the most reliable information in order to obtain the optimal combination of the different views.

 \subsection{Limitations}
 \paragraph*{Human Motion Assumption} 
 
 The proposed approach makes assumptions about human motion, which are stated in \cref{sec:peopleflow}. 
 In particular we assume that people are not sharing the same space and are moving at most one cell between two frames. 
 Both of this assumptions are conditioned on the discretization of the space. This is not a problem in the ground plane since people cannot share the same physical space and the ground plane grid can be sized to respect the motion constraints. On the other hand, it can be problematic to respect both constraints in the image plane. Therefore our method is best suited for calibrated setup: 
 full-camera calibration or at minimum a homography mapping the image plane to the ground plane. This is not a problem in a multi-view setting, but applying our method on a single view dataset would require an extra calibration step.

\paragraph*{Weakly Supervised Convergence} 

Predicting human motion while only being supervised with detection annotation is one of the merits of our approach. However it also has its downside, from \cref{eq:lossdet} and \cref{eq:losscycle} the flow is guaranteed to be accurate for small values of the loss $L$.
To reach high flow accuracy the training needs to have fully converged, this can be problematic on smaller datasets. Once the training reaches this converged state, it already starts to overfit the training data. On smaller datasets we observed this trade off between flow accuracy and generalization.


 \section{Conclusion}\label{sec:conclusion}
 In this paper, we propose a weakly supervised approach to detect people flow given only detection supervision. Our flow-based framework directly leverages temporal consistency across video frames and explicitly enforces scale and motion consistency over multiple viewpoints. Our experiments show that our model consistently outperforms state-of-the-art multi-view people tracking approaches. In the future, we plan extend our work to applications that track people in extremely crowded scene.
 
 \paragraph{Acknowledgments} This work was funded in part by the Swiss Innovation Agency.

 \clearpage

{\small
\bibliographystyle{ieee_fullname}
\bibliography{bib/optim,bib/string,bib/vision,bib/learning}
}

\end{document}